\documentclass{article}

\usepackage[a4paper, total={7in, 9in}]{geometry}

\usepackage{biblatex}
\usepackage{hyperref}
\usepackage{xurl}
\usepackage[english]{babel}
\usepackage{authblk}
\usepackage{algpseudocode}
\usepackage{graphicx}
\usepackage{algorithm}
\usepackage{amsmath}
\usepackage{booktabs}
\usepackage{amssymb}
\usepackage[toc,page]{appendix}
\usepackage{siunitx}
\usepackage{physics}

\title{Discovering Continuous-Time Memory-Based Symbolic Policies using Genetic Programming}
\author[1*]{Sigur de Vries}
\author[1]{Sander Keemink}
\author[1]{Marcel van Gerven}
\date{}
\affil[1]{Department of Machine Learning and Neural Computing, Donders Institute for Brain, Cognition and Behaviour, Radboud University,  Nijmegen, the Netherlands}
\affil[*] {sigur.devries@ru.nl}

\begin{document}
\title{Discovering Continuous-Time Memory-Based Symbolic Policies using Genetic Programming} 

\maketitle
\thispagestyle{plain}

\begin{abstract}
\noindent Artificial intelligence techniques are increasingly being applied to solve control problems, but often rely on black-box methods without transparent output generation. To improve the interpretability and transparency in control systems, models can be defined as white-box symbolic policies described by mathematical expressions. For better performance in partially observable and volatile environments, the symbolic policies are extended with memory represented by continuous-time latent variables, governed by differential equations. Genetic programming is used for optimisation, resulting in interpretable policies consisting of symbolic expressions. Our results show that symbolic policies with memory compare with black-box policies on a variety of control tasks. Furthermore, the benefit of the memory in symbolic policies is demonstrated on experiments where memory-less policies fall short. Overall, we present a method for evolving high-performing symbolic policies that offer interpretability and transparency, which lacks in black-box models.
\end{abstract}

\section{Introduction}
\begin{figure}[t]
    \centering
    \centerline{\includegraphics[width=\textwidth]{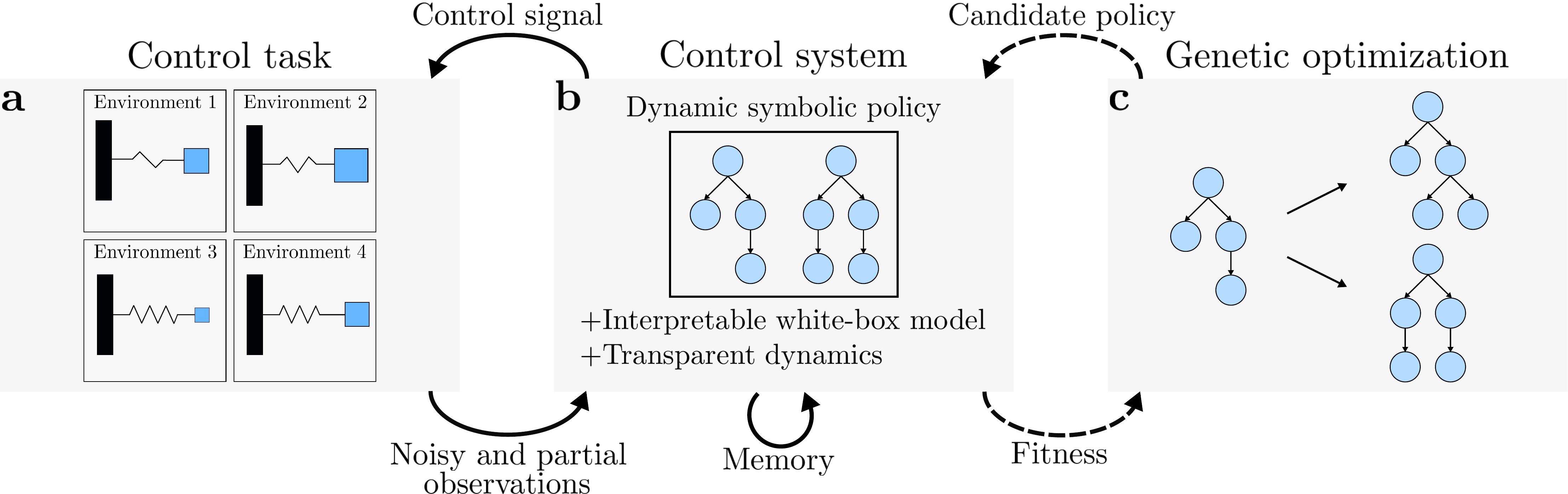}}
    \caption{\textbf{The general framework for discovering dynamic symbolic policies with genetic programming.} (\textbf{a})~The control task to be solved by the control system. The parameters in the environment may differ in each trajectory, and the observations are partial. (\textbf{b})~The control system is a dynamic symbolic policy, defined by interpretable symbolic expressions. The control system has internal dynamics that function as memory. The control system receives observations from the environment and outputs a control signal. (\textbf{c})~The symbolic policies are optimised with genetic programming. Candidate policies are evolved and evaluated on the control task which returns a fitness score, indicating the quality of the policy.}
    \label{Fig: overview}
\end{figure}

Many problems in our increasingly complex society can be viewed as control problems, where the goal is to make optimal decisions in real time such as to regulate the behaviour of the controlled system, ranging from smart infrastructure systems to healthcare systems~\cite{Annaswamy2023}. In recent years, there have been many successful applications of artificial intelligence (AI) to control problems, offering automated solutions to regulate complex non-linear processes. While the employed control algorithms achieve high performance, they usually require the use of black-box models such as neural networks~\cite{adadi2018peeking}. Black-box models often allow for highly general computations, but lack interpretability of their functioning, making it difficult to understand the computed decisions and actions~\cite{loyola2019black, glanois2024survey}. Transparency is particularly important in control systems, because it aids in identification of biases, fault detection and improvement and could therefore raise confidence of users in the system's reliability~\cite{zhang2021survey}. 

In contrast to black-box models, white-box models allow for easy inspection of their internal mechanisms~\cite{adadi2018peeking}. However, white-box models show an accuracy-interpretability trade-off and there are currently no general training methods to produce white-box approaches that can compete with black-box approaches in terms of efficiency and accuracy~\cite{loyola2019black,puiutta2020explainable}. Most white-box approaches opt for post-training model distillation to produce a transparent model that learns the input-output mapping of a black-box model~\cite{gou2021knowledge}. Disadvantages of model distillation are that it requires a two-phase learning process and information loss could occur in the distillation phase~\cite{ras2022explainable}. Therefore, it may be more desirable to learn white-box models directly. When optimising an inherently white-box model, interpretability of decisions already occurs at intermediate stages, showing transparent learning, and outputs are directly dependent on explainable features~\cite{ras2022explainable}.

A promising candidate for white-box control systems are symbolic policies~\cite{calegari2020integration, landajuela2021discovering,gou2021knowledge}. Symbolic policies are described by interpretable symbolic expressions, consisting of mathematical operators and state variables. Symbolic policies do not require a predefined model structure, therefore it is less affected by human bias or knowledge gaps compared to other data-driven approaches. Additionally, symbolic policies may capture rules that generalise better to other settings, and learned equations tend to be more efficient than neural networks in terms of parameter count~\cite{landajuela2021discovering}. 

Due to the non-differentiable structure of mathematical expressions, gradient-based methods cannot easily be applied to the task of the optimising symbolic policies. A viable alternative is genetic programming (GP), an evolutionary algorithm that automatically learns computer programs~\cite{koza1994genetic}. GP has seen great successes in symbolic regression problems, which seek to discover interpretable mathematical models describing data~\cite{cao2000evolutionary,sakamoto2001inferring,iba2008inference,schmidt2009distilling,quade2016prediction,bongard2007automated,wang2019symbolic}. Therefore we consider GP as a promising approach to optimise symbolic policies to solve control tasks~\cite{hein2018interpretable, nadizar2024naturally}. 

Current approaches to symbolic policy learning focus primarily on memory-less policies that have a fixed mapping from observations to control outputs~\cite{guo2024efficient, hein2018interpretable, nadizar2024naturally, landajuela2021discovering, gou2021knowledge, shimooka2000generating}. However, such policies may apply poorly to real-world settings, where volatility or partial observability complicate tasks. By adding memory to a control policy, past observations to improve policies in difficult settings~\cite{heess2015memory, khan2017memory}. This improvement has also been found by extending symbolic policies with memory~\cite{koza1992evolution, koza1999genetic, andersson1999reactive, raik1997evolving}, where GP explicitly learns to store and retrieve observations from memory. 

Alternatively, memory can be defined by dynamic variables that continuously integrate observations~\cite{rackauckas2020universal}. By modelling these variables with differential equations, the policy can use them to estimate unobserved variables or learn other important properties of the environment. This is similar to the application of neural differential equations to estimate parameterised dynamical systems~\cite{chen2018neural}. However, in contrast to neural differential equations, the latent variables in a symbolic policy are represented by symbolic equations, which makes the policy more transparent. Besides the latent variables, the policy also consists of a symbolic readout function that computes a control signal given the latent variables.

In this paper, we introduce a method for evolving memory-based symbolic policies with GP to solve dynamic control tasks, where continuous-time dynamically changing latent variables represent the memory (Fig.~\ref{Fig: overview}). Using this approach, symbolic policies are successfully evolved in linear and non-linear environments, under several challenging circumstances such as partial observability and varying environment settings, where memory-less policies obtain limited performance. High-performing symbolic policies are still discovered when the control dimension increases, showing that our method is scalable. Furthermore, we find that the evolved symbolic policies obtain performance comparable to the black-box baseline, given by neural differential equations~\cite{chen2018neural}. The key contribution of our work is a method for discovering interpretable high-performing memory-based symbolic policies using GP without requiring human expertise.

\section{Related Work}
Already in the early days of GP, it was proposed to find policies to solve control tasks~\cite{koza1994genetic, koza2000automatic}. These symbolic policies map the incoming observations to a control signal that was fed into the environment. Hereafter, GP has seen several applications for learning symbolic control policies~\cite{shimooka2000generating, hein2018interpretable, nadizar2024naturally}, although differentiable methods have also proven to be successful at finding symbolic policies~\cite{landajuela2021discovering, gou2021knowledge}. Furthermore, GP has been applied to the optimisation of decision trees~\cite{custode2023evolutionary} or fuzzy controllers~\cite{hein2018generating} to solve control tasks. Other evolutionary algorithms have shown to optimise control policies that compete with gradient-based approaches~\cite{salimans2017evolution,stanley2002evolving,such2017deep}.

In real-world settings, control policies should be able to handle partial observability and volatility in the environment. In~\cite{hein2018interpretable, hein2021trustworthy}, the policy receives observations from past time points, which enables the policy to control a stochastic and partially observable task, although the policy itself does not contain explicit memory. A method based on lagged observations requires storing a large number of data points, which does not remain robust when observations are noisy or in case of long trajectories~\cite{khan2017memory, lin1993reinforcement, mccallum1996hidden}, as this leads to a large increase of the search space in GP. A more general approach is to allow GP to learn to store computed values, which may be used for future computations~\cite{koza1999genetic, koza1992evolution, raik1997evolving, andersson1999reactive}. In~\cite{kelly2023discovering}, control systems with explicit memory storage are evolved with genetic algorithms, and the resulting policies are able adapt to changes in the environment. 

Contrary to these implementations of memory in control policies, in this paper we will use GP to evolve the state equations of continuous-time latent variables that function as memory. Similar approaches apply GP to optimise the structure of recurrent neural networks with internal memory~\cite{turner2017recurrent, esparcia1997evolving}. However in these works, the evolved networks operate in discrete time, individual nodes in the networks have values that are updated at each time step and the networks do not consist of symbolic equations. In our approach, incoming observations are continuously integrated into the dynamics of the latent variables and function as (recurrent) inputs for the functions in the symbolic control policy at any point in time. For clarity, in this paper the continuous-time memory-based policies are referred to as \textit{dynamic policies}, and the memory-less policies as \textit{static policies}.

The dynamic policies consist of multiple latent variables and readout functions, therefore multiple trees have to be optimised with GP simultaneously~\cite{langdon1998genetic}. In~\cite{cao2000evolutionary}, GP evolves multiple trees that together represent a system of coupled differential equations. Although the trees could be learned independently~\cite{bongard2007automated}, this setup is required when certain variables are not observed~\cite{cornforth2013inference}. This representation of multiple trees has also been utilised in collective decision making making~\cite{wen2016learning, brameier2001evolving} and multi-class classifiers~\cite{muni2004novel, ingalalli2014multi}. Based on these works, we extend our GP implementation to include tree based crossover and only apply crossover to pairs of trees from the same class.

\begin{figure}[t]
    \centering
    \includegraphics[width=\textwidth]{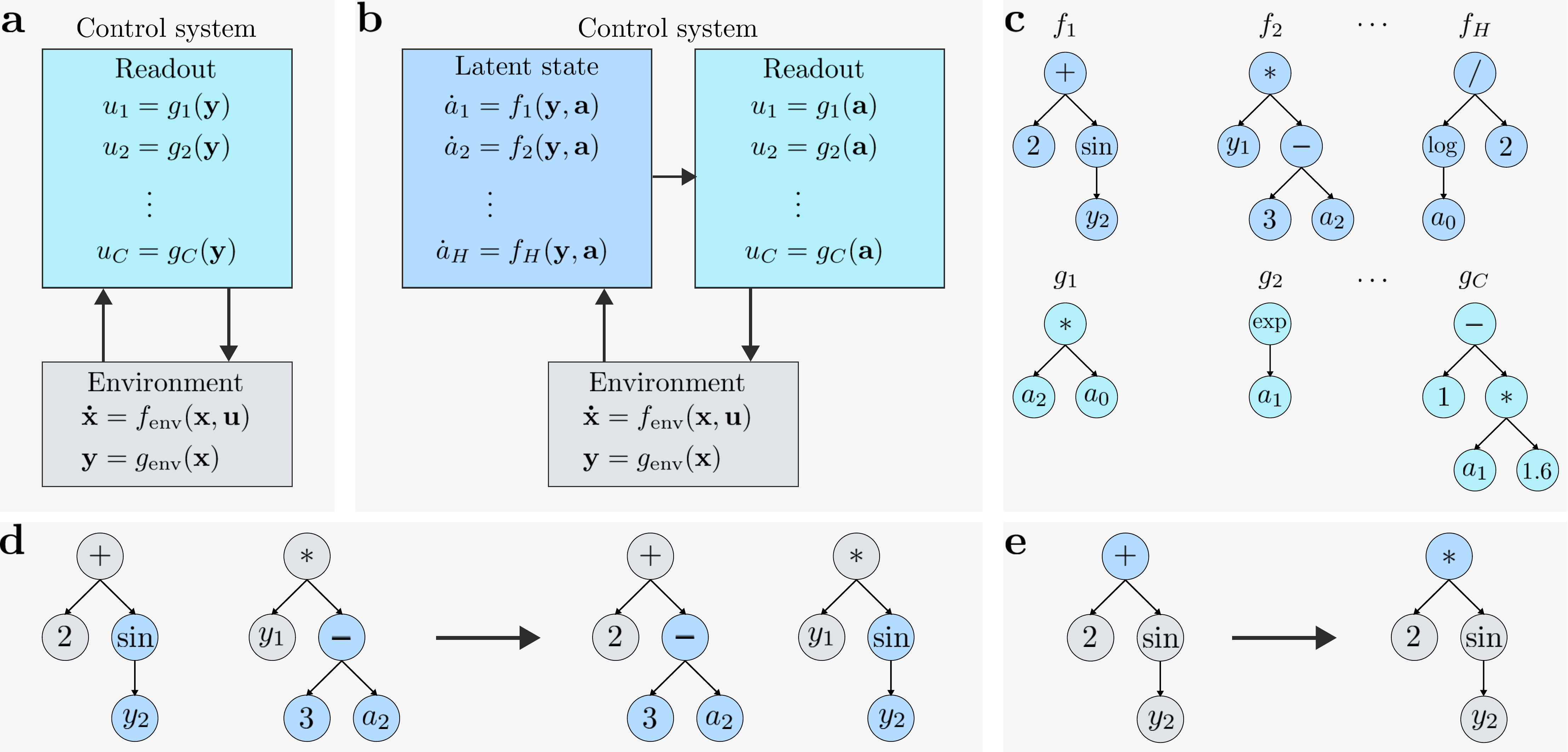}
    \caption{\textbf{Structure and optimisation of the static and dynamic symbolic policies}. (\textbf{a})~The feedback loop of an environment coupled with a control system that implements a policy without memory. \textit{C} represents the number of control signals required to interact with the environment. For every function $g_j$ a tree is optimised. (\textbf{b})~The control system is extended to include latent variables to add memory to the policy. \textit{H} represents the number of latent variables. (\textbf{c})~An example representation of how a symbolic policy with latent variables is represented by a set of trees. A tree is learned for the state equation of every latent variable and output signal. The functions $f_i$ and $g_j$ match the control system's structure defined in panel \textbf{b}. The trees are optimised with genetic programming. (\textbf{d})~In genetic programming, crossover is applied to a pair of parents and produces two offspring individuals. The blue subtrees are interchanged between the two trees. (\textbf{e})~Mutation changes one individual and results in a single offspring. In this example, the blue node is a function node that is changed to a different type of function node.}
    \label{Fig: methods}
\end{figure}

\section{Methods}
The goal of this paper is to evolve dynamic symbolic policies that obtain high performance in control problems. We compare dynamic memory-based symbolic policies with static memory-less symbolic policies, which are explained in more detail. Afterwards, the specifics of the GP algorithm and how we extended GP to learn multiple trees efficiently are described. In the subsequent subsections the baselines and experiments we performed are detailed. For the implementation of GP, we utilised an initial version of Kozax~\cite{de2025kozax}, a recently developed GP framework written in JAX. The code for reproducing the experiments and baselines is available at: \url{https://github.com/sdevries0/DSP_paper}.

\subsection{Environment model}
To evaluate a policy, a coupled system consisting of the policy and an environment is simulated, as described in Fig.~\ref{Fig: methods}a and Fig.~\ref{Fig: methods}b. In our paper, the policies are evaluated on dynamic tasks modelled as Markov decision processes. Throughout this paper, we will denote $x(t)$ as $x$ without explicit time dependence for notational convenience. The environment is modelled by a state $\mathbf{x} \in \mathbb{R}^N$, following the dynamics of a stochastic differential equation
\begin{equation}
    \dd \mathbf{x} = f_\textrm{env}(\mathbf{x}, \mathbf{u}) \dd t + \mathbf{V} \dd \mathbf{w} 
\end{equation} 
where $\mathbf{u} \in \mathbb{R}^{C}$ is the control input, $f_\textrm{env}$ is the environment's state equation, $\mathbf{w} \in \mathbb{R}^K$ is a multivariate Wiener process and $\mathbf{V} \in \mathbb{R}^{N\times K}$ determines how the process noise influences the dynamics.

The policy receives observations $\mathbf{y}\in \mathbb{R}^M$ from the state $\mathbf{x}$, determined by function $g_\textrm{env}$ and possibly corrupted by noise. In our experiments, the noisy observations are generated by \begin{equation}\label{eq: observation}
    \mathbf{y} = \mathbf{Dx} + \mathbf{\epsilon}
\end{equation} where $\mathbf{D} \in \mathbb{R}^{M\times N}$ is a readout matrix and $\mathbf{\epsilon} \sim \mathcal{N}(\mathbf{0}, \mathbf{\Sigma})$ is Gaussian observation noise, where $\mathbf{\Sigma} = \sigma\mathbf{I}$. When relevant, the policy receives a target state $\mathbf{x}^\ast$ that should be achieved in the environment. The policies are evaluated on a batch of trajectories, with a batch size of 32. In every trajectory, the initial conditions are randomly sampled, and if applicable target states and environment parameters are sampled.

\subsection{Control model}
\subsubsection{Static policies}
Symbolic control systems are conventionally defined as static symbolic policies~\cite{landajuela2021discovering, hein2018interpretable, guo2024efficient}, which directly map observations to a control signal (Fig.~\ref{Fig: methods}a). The policy is represented as a function $g$, which computes the control signal with only the observations $\mathbf{y}$ and potential target states $\mathbf{x}^\ast$ as input. This setup is given by
\begin{equation}\label{eq: readout}
    u_j = g_j(\mathbf{y}, \mathbf{x}^\ast)
\end{equation} for $1 \leq j \leq C$.

\subsubsection{Dynamic policies}
We extend this approach to dynamic symbolic policies (Fig.~\ref{Fig: methods}b), which implement memory through continuous-time dynamic latent variables. The latent variables process incoming observations, and the readout function maps the latent variables to a control signal. The dynamic policies contain expressions for the latent dimension and control dimension (Fig.~\ref{Fig: methods}c). The latent variables are governed by a system of ordinary differential equation according to
\begin{equation}
    \dot{a}_i = f_i(\mathbf{a}, \mathbf{y}, \mathbf{u}, \mathbf{x}^\ast)
    \label{eq:dynstate}
\end{equation} for $1 \leq i \leq H$, where we adopt the notation $\dot{x}=\dv*{x}{t}$ and $H$ represents the number of latent states in a policy and is chosen manually. Note that the dynamic policies also receive the previous control signal  as an additional input. The readout functions of the dynamic policies are similar to Equation~\ref{eq: readout}, but the input to $g_j$ is given by the latent variables $\mathbf{a}$ instead of the observations $\mathbf{y}$, which results in
\begin{equation}
    u_j = g_j(\mathbf{a}, \mathbf{x}^\ast)
    \label{eq:dynreadout}
\end{equation}
for $1 \leq j \leq C$. 

\subsection{Numerical integration}

Both the dynamics of the environment and the controller are simulated as a coupled system over a specified time range [$t_0, t_f$] with $t_0$ the initial time and $t_f$ the final time.  
For the environment and dynamic policies, we need to resort to numerical integration. To this end, we make use of the Diffrax library~\cite{kidger2021on}. Because the systems are stochastic, the integration is performed using the Euler-Heun method with a fixed step size $\Delta t$~\cite{kloeden1994}. 
This results in trajectories $\mathbf{x}_{0:f} = (\mathbf{x}_0, \ldots, \mathbf{x}_f)$ and $\mathbf{u}_{0:f}= (\mathbf{u}_0, \ldots, \mathbf{u}_f)$ with $\mathbf{x}_i$ and $\mathbf{u}_i$ the state and control at time $t_i$. These trajectories are used to determine the policy's fitness using a fitness function $F(\mathbf{x}_{0:f}, \mathbf{u}_{0:f})$.

\subsection{Genetic programming}\label{sec: GP}
The symbolic policies are evolved via genetic programming.  GP is a variant of evolutionary algorithms that focuses on learning computer programs~\cite{koza1994genetic}. A population of solutions is evolved to satisfy a certain goal or task through stochastic optimisation. In our setup, individuals are represented as tree-structured computational graphs, constructed from a predefined set of function nodes and leaf nodes. Function nodes cover mathematical operators and functions, while leaf nodes describe variables and constants.  Two important hyperparameters in GP are the number of generations, which determines how many iterations of evolution are performed, and the population size, which defines the number of candidates that are evolved at every generation. These hyperparameters should be balanced to reach convergence while remaining computationally efficient.  
See Algorithm~A1 for an overview of the GP algorithm. The GP algorithm consists of initialisation, evaluation and reproduction steps, as described in more detail in the following sections.

\subsubsection{Initialisation} \label{sec: GP_init}
Initially the population consists of randomly generated individuals, which will generally score poorly on the problem. Nonetheless, the initial population can already have a large effect on the success of the full algorithm~\cite{beadle2009semantic}. The initial population should cover enough of the search space for the genetic operations to be useful. The initialisation strategy that was originally introduced is called ramped half-and-half~\cite{koza1994genetic}. In this method, trees are either sampled fully, where leaf nodes can only appear at the maximum depth, or the trees grow randomly, which means that leaf nodes can appear at earlier depths, so that the initial population covers both shallow and deep trees.

\subsubsection{Evaluation} \label{sec: GP_eval}
To evaluate a candidate policy, the trees are transformed into callable programs and tested on a problem. The performance of the individual is expressed with a fitness score, computed with a fitness function. The fitness scores are used to select individuals for reproduction, where fitter individuals produce more offspring. After evaluation, parents are selected for reproduction using tournament selection~\cite{fang2010review}.

\subsubsection{Reproduction} \label{sec: GP_rep}
Every generation a new population is generated, consisting of candidate solutions that are generated with crossover and mutation. In crossover the genotype of two individuals is combined to produce new trees. Fitter individuals are more likely to be selected for reproduction, therefore the offspring is built with promising tree structures. In both parents a random node from the tree is selected, after which these nodes and their subtrees are swapped. An example of the crossover operator is shown in Fig.~\ref{Fig: methods}d. Uniform crossover is a variant of the standard crossover where multiple nodes can be swapped between a pair of trees without swapping full subtrees~\cite{spears1991study}. Uniform crossover improves the locality of the crossover operation, but both types of crossover can produce fit individuals and are therefore both included in our implementation.

Mutation was originally not integrated into GP~\cite{koza1994genetic}, but mutation has shown to be a useful addition to the algorithm~\cite{luke1997comparison}. Mutation is applied to a single individual to generate one offspring and is applied instead of crossover in our implementation. Many aspects of trees can be mutated, for example changing, deleting or adding operators, replacing subtrees or changing variables or constants. Fig.~\ref{Fig: methods}e shows an example of mutation, in which an operator is changed to a different type of operator. Besides crossover and mutation, the new population also consists of randomly generated individuals and the elite individuals from the previous generation.

The number of trees that has to be optimised differs for the static and dynamic policies. The static policies evolve a tree for every control input the problem requires. These trees also have to be evolved in the dynamic policies, but additional trees are optimised for every latent variable. We chose to fix the dimensionality of the latent state in the dynamic policies and allow the readout function to learn which latent variables to use. To learn both a symbolic latent state and readout function simultaneously in the dynamic policies, the individuals in the GP algorithm are extended to learn multiple trees~\cite{cao2000evolutionary}. During reproduction, crossover and mutation are not applied to every tree in the individual to be able to make small jumps through the search space. Crossover is only applied to trees in the same position in two parents, and a third type of crossover is introduced in which only complete trees are swapped between individuals~\cite{muni2004novel}.

\subsubsection{Regularisation}
Regularisation is necessary in GP to produce interpretable trees and improve the efficiency of the optimisation. To regularise the complexity of the reproduced trees, a punishment relative to the number of nodes in a tree is added to the fitness~\cite{luke2006comparison}. Additionally, the depth of the trees is limited to seven, and during initialisation the depth is set to four. To introduce more compact trees in the population, a small part of the population is simplified mathematically at every generation~\cite{hooper1996improving}. The equations are simplified using SymPy~\cite{meurer2017sympy}, after which they are restructured into the tree representation. To improve diversity in the population during evolution, multiple subpopulations are evolved independently~\cite{fernandez2003empirical}. Each subpopulation is evolved with different selection pressure, probability of a tree to be changed during evolution and probabilities for crossover, mutation, random sampling and simplification. This way, subpopulations with high crossover and simplification probability focus more on exploitation, while exploration is enforced for subpopulations with high mutation and random sampling rates, as well as higher selection pressure and probability to evolve a tree in a candidate. Configuring subpopulations with different hyperparameters to balance exploration and exploitation is beneficial~\cite{herrera2000gradual}, especially combined with migration between the subpopulations.

\subsection{Baselines}
The static and dynamic symbolic policies generated with GP are compared with two baselines. The baselines are evaluated with the same population size and number of generations for every experiment. 

The first baseline is random search of dynamic symbolic policies, which randomly samples a set of trees for each policy. The policies have the same model structure as the dynamic policies (Fig.~\ref{Fig: methods}c), and consist of the same set function and leaf nodes. The maximum depth of trees is fixed to seven. Random search is included as a baseline to demonstrate if GP can find better solutions with less computation than random sampling, therefore showing the additional value of the genetic operations during optimisation.

To verify that the symbolic policies are high-performing, the symbolic policies are also compared to black-box policies modeled by neural differential equations (NDEs)~\cite{chen2018neural}. 
The NDEs in our experiments implement Equations~\ref{eq:dynstate} and \ref{eq:dynreadout} using black-box neural networks. Specifically, we use
\begin{equation}\label{Eq:NDE}
    \dot{\mathbf{a}} = \tanh(\mathbf{A}\mathbf{z})
\end{equation} 
with $\mathbf{z} = (\mathbf{a}, \mathbf{y},\mathbf{u},\mathbf{x}^\ast, 1)$ a column vector concatenating the inputs and a constant, and $\mathbf{A} \in \mathbb{R}^{N \times |\mathbf{z}|}$ a matrix.
The linear readout layer is defined as \begin{equation}
    \mathbf{u} = \mathbf{B}\mathbf{}\mathbf{v}
\end{equation} 
with $\mathbf{v} = (\mathbf{a}, \mathbf{x}^\ast, 1)$ and $\mathbf{B} \in \mathbb{R}^{C \times |\mathbf{v}|}$.

In this case, model parameters are optimised using covariance matrix adaption evolution strategies~\cite{hansen2016cma}.
In all experiments, the dimension of $\mathbf{a}$ was set to five for the NDE baseline. The NDE policies are included as a baseline to investigate whether symbolic policies can compete with a proven black-box method.

The performance of our method and the baselines is demonstrated through the best fitness at every generation during evolution, averaged over 20 independent evolutionary runs. The validation fitness of the best policy found in every run is displayed to confirm that the policies generalise to unseen conditions, but also to inspect the variance of convergence in the different evolutionary processes.

\subsection{Environments}\label{sec: environments}
Our method and the baselines are evaluated on three different environments. The policies are tested on a linear and non-linear classical control task and on a non-linear industrial problem. The number of generations, population size and latent state size used in each experiment are presented in Appendix Table~\ref{Table: Hyper_params}. 
The choice for the population size and number of generations were chosen to balance efficiency and consistent convergence empirically. To ensure that the policies have enough computational complexity, number of latent states in the dynamic symbolic policies was set to two in every experiment that requires one control input, and set to three in the experiment with multiple control inputs. The choice of the function sets are described in the following subsections.

\subsubsection{Stochastic harmonic oscillator}\label{sec: HO_exp}
\begin{figure}[t]
    \centering
    \includegraphics[width=0.9\textwidth]{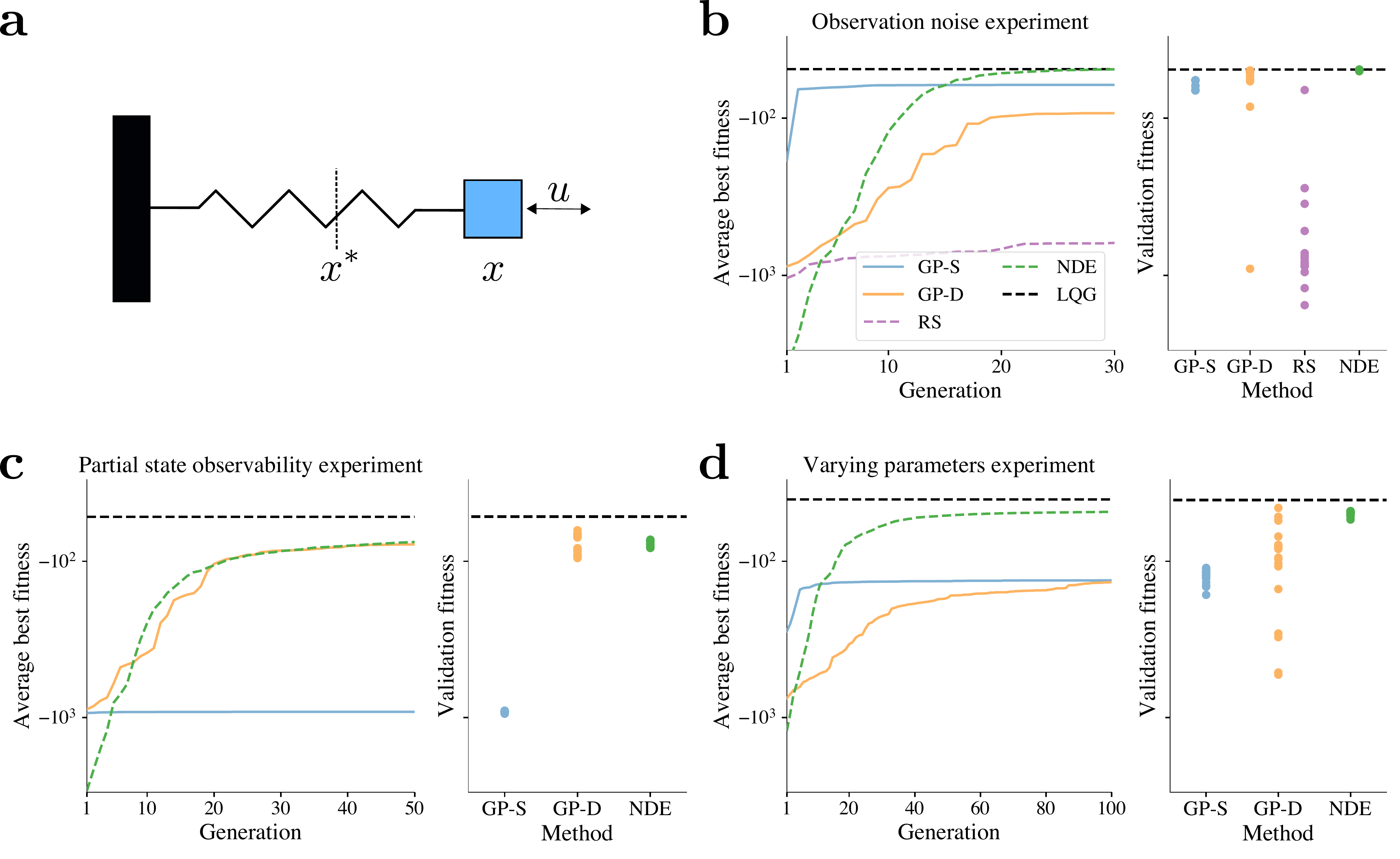}
    \caption{\textbf{Genetic programming evolves high-performing dynamic symbolic policies for the stochastic harmonic oscillator.} (\textbf{a})~In the stochastic harmonic oscillator (SHO), the position of the oscillator $x$ has to be moved to a target state $x^*$ by applying the control force $u$. (\textbf{b})~Evolution results of the experiment on the SHO with observation noise. The left plot shows the best fitness at every generation during evolution, averaged over 20 independent runs. The methods included are genetic programming for static (GP-S) and dynamic (GP-D) policies, random search (RS), neural differential equation (NDE) and the analytical optimal linear quadratic Gaussian (LQG). The right plot presents the validation fitness obtained on unseen conditions with the best policy evolved in every evolutionary run. (\textbf{c})~Evolution results for the experiment in which the state of the SHO is partially observable. (\textbf{d})~Evolution results of the experiment on the SHO with varying environment parameters.}
    \label{Fig: ho_results}
\end{figure}

The stochastic harmonic oscillator (SHO) is a linear system that describes the position and velocity of an oscillating mass in a single dimension under random perturbations. The SHO is visualised in Fig.~\ref{Fig: ho_results}a. The goal of the controller is to stabilise the mass at a target position while minimising the control usage. The SHO is easily solved with a linear controller, therefore we compute the linear quadratic Gaussian (LQG) control as an optimal baseline for comparison. The policies are tested in a setting with noisy observations, a setting where only the position of the oscillator is observed, and a setting where the environmental parameters vary per trajectory. To discover generalising policies, each policy is tested on a batch of trajectories, with random initial conditions and targets. Since the optimal controller is linear, the function set of GP consists of the basic arithmetic operators. See Appendix B for details on the dynamics, the fitness function and the specific definitions of the experiments.

\subsubsection{Acrobot swing up task}\label{sec: ACR_exp}
The acrobot is defined by two connected links, where the first link is connected to a fixed position (Fig.~\ref{Fig: acrobot_results}a). The goal is to swing the end point of the second link above a threshold, while minimising the applied force. The dynamics of the acrobot are non-linear, and to accomplish the swing up a non-linear controller is required. The policies solve the acrobot in experiments with noisy observations including and excluding the observations velocity, and a third experiment is conducted where control can be applied to both links individually. The observations of the acrobot include angles, therefore the sine and cosine functions are added to the function set of the GP algorithm, besides the arithmetic operators. The dynamics and the fitness function are described in Appendix C.

\subsubsection{Continuous stirred tank reactor}
The continuous stirred tank reactor (CSTR) describes a chemical process that is governed by non-linear dynamics~\cite{uppal1974dynamic}. The CSTR consists of a main tank that contains a mixture of chemicals and a cooling jacket around the main tank. A visualisation of the CSTR is presented in Fig.~\ref{Fig: cstr_all}a. To optimise the chemical process inside the main tank, the temperature has to be stabilised around a setpoint temperature. The temperature in the cooling jacket can be influenced directly by altering the input flow of coolant, which in turn affects the temperature in the main tank. in our experiment, the parameters of the CSTR are varied in every trial (Table~III). The policy receives noisy observations of the temperature in the reactor and cooling jacket, but the concentration is not observed. The function set of the GP algorithm is extended to include the exponential and logarithmic functions to aid expressions to deal with the values of the temperatures, which typically range between 200 and 600. The dynamics and the values of the environment's parameters are detailed in Appendix D, as well as the fitness function.

\section{Results}\label{sec: results}
We present a GP method for finding high-performing dynamic symbolic policies for solving control problems. The performance of a policy is tested by controlling three common control benchmark systems (as explained in detail in Section~\ref{sec: environments}). We investigated whether GP evolves successful control policies for linear and non-linear environments, different observability levels, as well as changing environment conditions and multiple control dimensions. We compared the resulting dynamic symbolic policies to static symbolic policies, optimal control (where possible), random search (RS), and to a reference black-box model (NDEs).

\begin{table}[t]
\centering
\caption{Fitness of the best policies on validation trajectories for every experiment. The complexity of the best policies is also displayed as the size. The methods include random search (RS), neural differential equation (NDE) and genetic programming for static (GP-S) and dynamic (GP-D) policies. The bold fitness indicates that a method obtained the best fitness in an experiment. The complexity is computed by summing the number of mathematical operations, variables and constants in the policies.}
\vspace{0.2cm}
 \begin{small}
 \centering
\begin{tabular}{lllllllllllll}
\toprule
 &&\multicolumn{2}{c}{\textbf{RS}}&&\multicolumn{2}{c}{\textbf{NDE}}&&\multicolumn{2}{c}{\textbf{GP-S}} &&\multicolumn{2}{c}{\textbf{GP-D}}\\
 \cmidrule(lr){3-4} \cmidrule(lr){6-7} \cmidrule(lr){9-10} \cmidrule(lr){12-13}
  \textbf{Environment}&  \textbf{Experiment}  & Fitness & Size & &Fitness & Size&&Fitness & Size &&Fitness & Size\\
 \midrule
 SHO & Observation noise & -56.11 &15 &&\textbf{-48.85} & 215&&-57.28 &5 &&-49.66&19\\
 SHO & Partial state observability & - & - && -72.76&195& & -903.44 &5&& \textbf{-63.35}&21\\
 SHO & Varying parameters & - & - && -47.76 &215&& -110.25 &13&& \textbf{-45.56}&25\\
 Acrobot & Observation noise & -123.64 & 6&& \textbf{-98.45} &231&& -115.31 & 8&& -112.82&20\\
 Acrobot & Partial state observability & - & - && \textbf{-101.47} & 191&& -250.0 & 1&& -110.23&12\\
 Acrobot & Two control inputs & - & - && \textbf{-97.02}& 272&& -103.43 &12&& -100.69&18\\
 CSTR & Standard & - & - && -1017.72 & 215 && -1059.31 & 9 && \textbf{-845.33}& 27\\
 \bottomrule
\end{tabular}
\end{small}
\label{Table: fitness_table}
\end{table}

\subsection{Genetic programming evolves linear and non-linear symbolic policies}
To study whether GP can successfully evolve dynamic symbolic policies, we first consider the SHO and acrobot environments with noisy full state observability (Figs.~\ref{Fig: ho_results}b and \ref{Fig: acrobot_results}b). In the SHO, we define the linear quadratic Gaussian (LQG) controller as the upper bound fitness (horizontal black lines). The fitness of random search improved during optimisation (purple curves), but did not converge near the upper bound nor the other methods. Random search did occasionally discover a policy that competes with the evolutionary approaches, but not consistently (purple dots). Moving forward, we will no longer consider random search as a baseline, but the results were similar across all tasks. NDEs converged to a high fitness quickly and consistently, approximating the upper bound fitness in the SHO (green curves and dots).

To validate that symbolic policies reach similar performance as the NDE, we first optimised static symbolic policies (blue curves). The initial fitness of GP-S started close to the upper bound fitness in the SHO, which indicates that well-performing static policies exist in the initial population, which is a result of the small search space of GP-S. However, from there, GP was not able to evolve policies reaching the optimal fitness, indicating that static policies are not complex enough to match the optimal controller. Similarly, in the acrobot swing-up task, GP-S began at a high fitness, but did not improve much during evolution. In both tasks, GP-S converged significantly higher than RS, but was outperformed by the NDE in terms of fitness.

We next evolved dynamic symbolic policies (orange curves). In both tasks, the fitness of GP-D began significantly lower than GP-S, due to the larger search space. However, GP-D quickly converged to a higher fitness than GP-S, approximating the upper bound fitness in the SHO. This indicates that the internal dynamics are able to correct for the noisy observations. Both the GP-S and GP-D policies are highly consistent across trials, with GP-D producing slightly better policies (blue and orange dots).

Overall, these experiments demonstrate that static and dynamic symbolic policies are effectively optimised with GP, especially compared to random search. High-performing linear and non-linear policies were evolved that compete with the black-box NDE, although the NDE is more consistent.

\subsection{Dynamic symbolic policies remain robust under partial state observability}
\begin{figure}[t]
    \centering
    \includegraphics[width=0.9\textwidth]{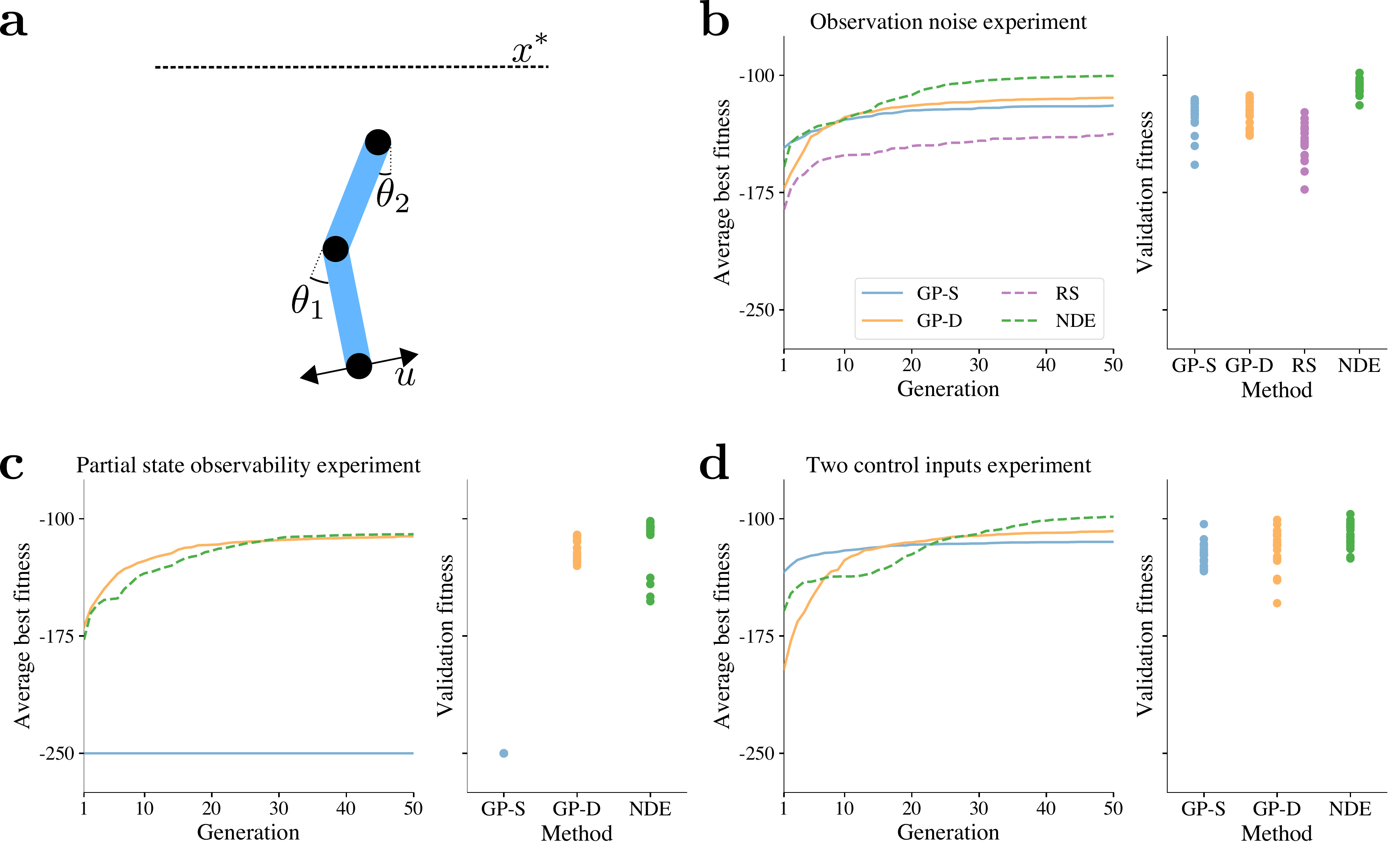}
    \caption{\textbf{Genetic programming evolves high-performing dynamic symbolic policies for the acrobot swing-up task.} (\textbf{a})~The acrobot consists of two links, which are each defined by the angles $\theta_1$ and $\theta_2$. To complete the swing up, the end of the second link has to be moved above the target height $x^*$ by applying control force $u$ on the second link. (\textbf{b})~Evolution results of the experiment in the acrobot environment with observation noise. The left plot shows the best fitness at every generation during evolution, averaged over 20 independent runs. The methods included are genetic programming for static (GP-S) and dynamic (GP-D) policies, random search (RS) and neural differential equation (NDE). The right plot presents the validation fitness obtained on unseen conditions with the best policy evolved in every evolutionary run. (\textbf{c})~Evolution results of the experiment on the acrobot under partial state observability. (\textbf{d})~Evolution results of the experiment on the acrobot with two control inputs.}
    \label{Fig: acrobot_results}
\end{figure}

The previous experiments demonstrated that static and dynamic symbolic policies are successfully evolved for linear and non-linear settings. However, these experiments assumed the full state is observed, which may not be representative for real-life applications. We hypothesised that the memory in dynamic policies is particularly beneficial when confronted with partial state observations. To test this hypothesis, symbolic policies were evolved for the SHO and acrobot environments under partial state observability, where the velocities are not observed (Figs.~\ref{Fig: ho_results}c and \ref{Fig: acrobot_results}c).

The SHO remains linear, therefore the LQG controller again represents the upper bound fitness (horizontal black lines). The NDE converged to a slightly lower fitness (green curves) compared to the experiments with full state observability. However, the NDE still consistently produced policies close to the upper bound in terms of performance, with a few policies that perform worse on the acrobot (green dots).

As hypothesised, GP-S converged to a substantially lower fitness than under full state observability (blue curves). Every static symbolic policy fails to satisfy the task when the velocity is not observed (blue dots). Differently from GP-S, high-performing dynamic symbolic policies were still successfully evolved (orange curves). GP-D converged to a similar fitness as the NDE in both environments. Additionally, the validation fitness is consistent, and some policies even perform better than the NDE policies (orange dots).

\begin{figure}
    \centering
    \includegraphics[width=0.6\linewidth]{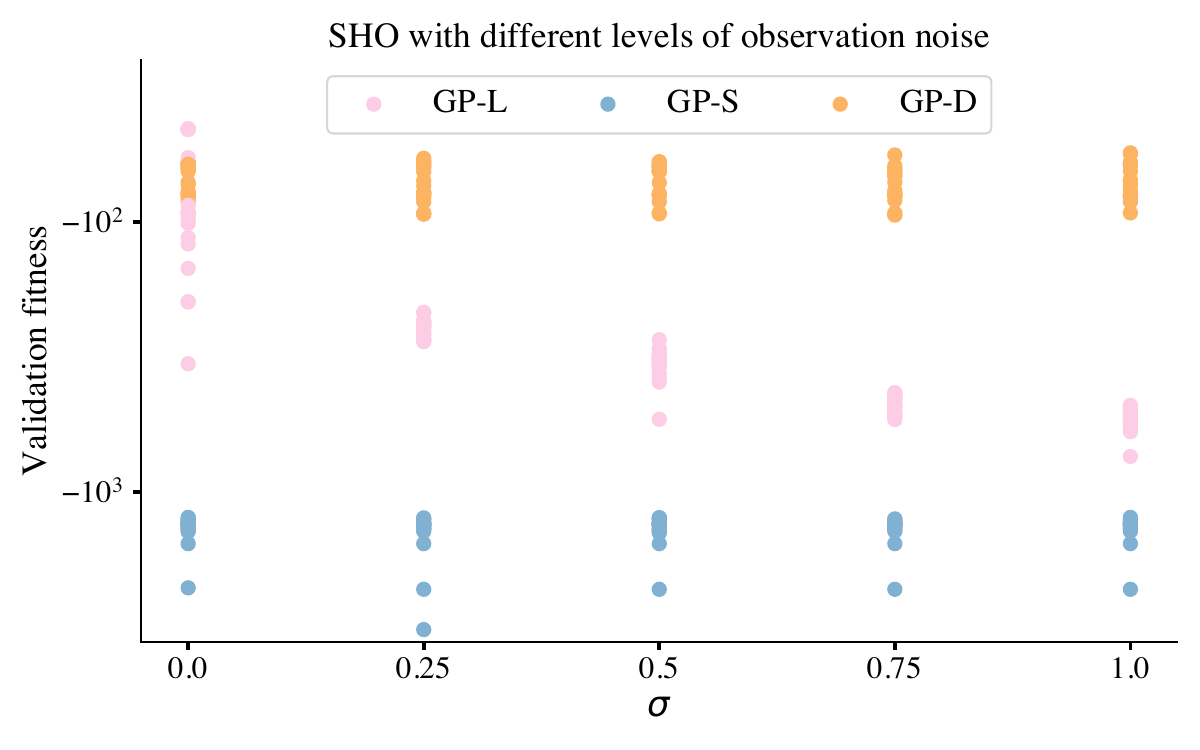}
    \caption{\textbf{Dynamic policies remain more robust than time-lagged policies under higher levels of noise.} Policies without memory (GP-S), with latent memory (GP-D) and access to time-lagged variables (GP-L) were evolved on the stochastic harmonic oscillator (SHO) with partial state observability and varying levels of observation noise $\sigma$. The average validation fitness on unseen conditions is shown for the policies evolved with 20 different seeds. The time-lagged policies received observations from the last three time steps as input.}
    \label{fig: lag comparsion}
\end{figure}

This experiment demonstrated that static symbolic policies perform poorly under partial state observability compared to the dynamic symbolic policies and NDE. This result can be contributed to the lack of memory in the static policies. Memory allows for estimation of the velocity, which enables policies to solve the SHO and acrobot tasks successfully. The latent state significantly increases the search space, but improves the robustness of the dynamic symbolic policies when confronted with partial observability.

Naively applying static policies to partially observable environments is expected to perform poorly. In practice, researchers would provide time-lagged observations to improve control policies~\cite{hein2018interpretable, mnih2013playing, uehara2022provably}. Lagged variables in principle allow programs to infer more information and deal with e.g. partial observability by estimating missing variables. However as the level of noise increases, more history of observations are required for accurate estimation. Consequently, the search space of GP increases with more observations and performance deteriorates with limited history length. The latent dynamic memory proposed in this work does not suffer from this limitation, as observations are continuously integrated. 
To demonstrate this, Fig.~\ref{fig: lag comparsion} shows the comparison between GP-S, GP-D and policies that receive time-lagged observations (GP-L, pink dots) on the SHO with partial state observability and increasing $\sigma$ (Eq.~\ref{eq: observation}), representing more noise on the observation of the position. GP-L received the three most recent observations of the position at every time step. With these time-lagged observations of the position, the velocity can be estimated and used for accurate control. Fig.~\ref{fig: lag comparsion} shows that without observation noise, GP-L (pink dots) is competitive with GP-D (orange dots) and outperforms GP-S (blue dots). However as the observation noise increased, it became more difficult to estimate the velocity and the performance of GP-L deteriorates, moving closer to GP-S in terms of performance. GP-D remained constant in performance when the observation noise increased, demonstrating more reliable estimation capabilities through continuous integration of observations. The best policies of GP-L, GP-S and GP-D are presented in Appendix Table~\ref{Table: lag_policy_table} for every noise level. Increasing the number of time-lagged observations could improve the estimation given high noise levels, but also scales the size of the search space. The dynamic policies generalise the time-lagged observations, as the latent variables can integrate an arbitrary length of time steps.

\subsection{Symbolic policies generalise to varying environments}
Besides partial observations, realistic settings also require policies to generalise to varying and changing environments. To validate that generalising symbolic policies can be evolved, we tested policies in the SHO with parameters that vary between trials (Fig.~\ref{Fig: ho_results}d). In this experiment, the upper bound fitness was again obtained with the LQG controller that is computed independently for each parameter setting (horizontal black lines). The NDE converged close to the upper bound fitness and is consistent across different runs (green curve and dots). 

Static policies were evolved to validate their generalisability (blue curve). GP-S did not converge near the upper bound fitness and the NDE. The validation fitness of the best static policies shows that a ceiling has been reached, as every policy is located close to the average (blue dots). GP-D converged to a similar fitness as GP-S (orange curve). However, the validation fitness of the dynamic symbolic policies shows an interesting pattern (orange dots). Many dynamic symbolic policies obtain a validation fitness that is better than any static policy, and some even compare with NDE policies. However, some dynamic symbolic policies obtain a validation fitness worse than any static policy.

Appendix Fig.~\ref{fig: variance}a shows that most policies perform similarly on the test data as on their respective training data sets. However, some policies fail to generalise, which is mostly the case for policies that include one or multiple non-linear operations. As the harmonic oscillator is governed by linear dynamics, the presence of non-linear operations in a policy could relate to overfitting on the training data and consequently harms the ability to generalise. The worst policy in terms of validation fitness is however extremely simple, as it does not make use of the latent memory. 

The inconsistent evolution of dynamic symbolic policies is explained by the large search space and overfitting. When dynamic policies are evolved successfully, they perform better than static policies. The memory allows dynamic policies to discover useful properties that increase generalisability. In our results, the majority of evolutionary runs resulted in dynamic policies that exceed the performance of static policies. Therefore the addition of the latent state in symbolic policies proves to be more beneficial on average. Our approach does not consistently produce policies that perform comparably as the NDE policies, though this comes at the expense of obtaining white-box solutions.

\begin{table}[t]
\centering
\caption{The best static and dynamic symbolic policies discovered with genetic programming for every experiment. $u_j$ represents the function that outputs the control signal. In the dynamic policies, the functions of $a_i$ represent the state equations of the latent variables. When a latent variable is not referred to in any of the other expressions, it is labelled as inactive.}
\vspace{0.2cm}
 \begin{small}
 \centering
\begin{tabular}{llll}
\toprule
 \textbf{Environment}&  \textbf{Experiment}&\textbf{Static policy} &\textbf{Dynamic policy}\\
 \midrule
 SHO & Observation noise & $\begin{array}{lll}u_1 &=& - 0.61y_2 + x^\ast\end{array}$  & $\begin{array}{lll}u_1 &=& -2a_1 + 2.60a_2 + x^\ast \\ \dot{a}_1 &=& y_2\\ \dot{a}_2 &=& -u_1 + x^\ast\end{array}$ \\
 \midrule
 SHO & Partial state observability & $\begin{array}{lll}u_1 &=& 0.75x^\ast - 0.11\end{array}$  & $\begin{array}{lll} u_1 &=& 0.45(a_1+x^\ast)\\ \dot{a}_1 &=& 2 a_2 - u_1 + x^\ast\\ \dot{a}_2 &=& -a_2 - 0.99 u_1 + y_1\end{array}$ \\
 \midrule
 SHO & Varying parameters & $\begin{array}{lll}u_1 &=& -1.10 y_1 - 0.73 y_2 \\&&+ 1.83 x^\ast + 0.17\end{array} $  & $\begin{array}{lll}u_1 &=& -a_1^3 - a_2 + x^\ast\\ \dot{a}_1 &=& 27.76 (-a_1 + y_1 + y_2 - x^\ast) \\ \dot{a}_2 &=& 0.32 a_1\end{array}$ \\
 \midrule
 Acrobot & Observation noise & $\begin{array}{lll}u_1 &=& -y_3 + 1.29 \sin(y_4)\end{array}$  & $\begin{array}{lll}u_1 &=& 2 a_2 - \cos(2 a_1) \\ \dot{a}_1 &=& 2 y_1-2 y_2 \\ \dot{a}_2 &=& 5.47 \sin(\sin(y_1))\end{array}$ \\
 \midrule
 Acrobot & Partial state observability & No successful policies  & $\begin{array}{lll}u_1 &=& 1.87 a_1 + \cos(a_2)\\ \dot{a}_1 &=& 2.06 y_1 \\ \dot{a}_2 &=& y_1-2.68\end{array}$ \\
 \midrule
 Acrobot & Two control inputs & $\begin{array}{lll}u_1 &=& 0.64y_4 + \sin(y_4) - 0.04  \\ u_2 &=& \sin(0.37y_3) \end{array}$ & $\begin{array}{lll}
 u_1 &=& 2.71\cos(a_2-0.57)\\
 u_2 &=& -a_2 - 1.48 \\
 \dot{a}_1 &=& -8.66a_1 + y_4 \,\text{(inactive)}\\
 \dot{a}_2 &=& -3.44y_2\\
 \dot{a}_3 &=& y_3 \,\text{(inactive)}
 \end{array}$\\
 \midrule
 CSTR & Standard &  $\begin{array}{lll}u_1 &=& T_c^4 (T_r^\ast)^{-3} - T_r\end{array}$  & $\begin{array}{lll}u_1 &=& a_1^2 a_2 + 2 a_1 \log(T_r^\ast)\\ \dot{a}_1 &=& (2.92-a_1)(u_1 + 2.98) \\ \dot{a}_2 &=& (u_1 - a_2 + 2.37)(a_2 - T_r^\ast + T_r)\end{array}$ \\
 \bottomrule
\end{tabular}
\end{small}
\label{Table: policy_table}
\end{table}

\subsection{Genetic programming evolves higher-dimensional policies}
As the search space of symbolic policies is large, especially for the dynamic symbolic policies, it is important to validate whether high-performing policies are found efficiently when the control dimension increases. In this experiment, the policies can apply control forces to both links to accomplish the acrobot swing-up (Fig.~\ref{Fig: acrobot_results}d). The NDE converged to a slightly higher fitness during evolution than in the other experiments on the acrobot, because more force could be applied to solve the swing up (green curve). The best NDE policies obtained consistent validation fitness (green dots).

Our method was first tested to produce static symbolic policies with two outputs (blue curve). The search space of static policies is still small compared to the dynamic policies, as is observed from the higher initial average best fitness compared to the other methods. The average best fitness of GP-S did not improve much during evolution, and converged at a worse fitness than the NDE. The validation fitness shows that most static policies perform similarly, but one evolved policy performs better than the other static policies (blue dots). GP-D converged at a slightly better fitness than GP-S, but worse than NDE (orange curve). The validation fitness obtained by dynamic symbolic policies mostly overlaps with that of the static symbolic policies and NDE policies, although a few policies are worse than any policy obtained by the other methods (orange dots). This experiment demonstrates that GP still evolves higher-dimensional control policies that accomplish the acrobot swing up, even though the size of the search space was increased even further.

\subsection{Generalising symbolic policies are discovered for an industrial application}
\begin{figure}[t]
    \centering 
    \includegraphics[width=\textwidth]{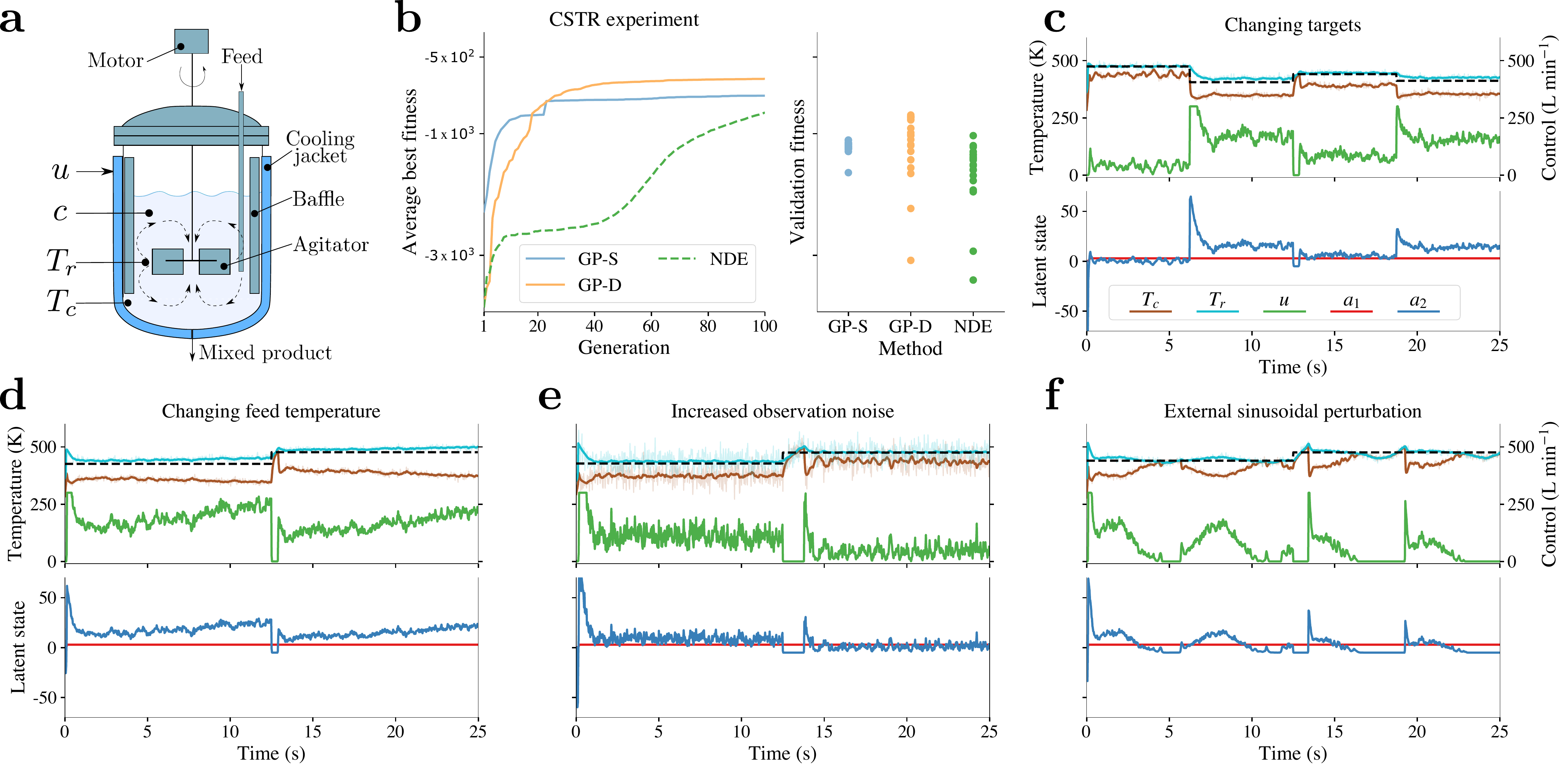}
    \caption{\textbf{Dynamic symbolic policies generalise to industrial control settings.} (\textbf{a})~The continuous stirred tank reactor (CSTR) is described by the concentration of reactant $c$, the temperature $T_r$ in the main reactor and the temperature $T_c$ in the cooling jacket. The temperature in the reactor should be stabilised at a set point temperature $T_r^*$. The flow of coolant into the cooling jacket is controlled by $u$. (\textbf{b})~Evolution results of the experiment on the CSTR with partial state observability and varying parameters. The left plot shows the best fitness at every generation during evolution, averaged over 20 independent runs. The methods included are genetic programming for static (GP-S) and dynamic (GP-D) policies and neural differential equation (NDE). The right plot presents the validation fitness obtained on unseen conditions with the best policy evolved in every evolutionary run. (\textbf{c})~Simulated trajectory of the CSTR controlled with a dynamic symbolic policy. The top row shows the temperature in the reactor and cooling jacket and the controlled coolant flowrate into the cooling jacket with varying setpoint temperatures. The bottom row shows the latent variables of the policy. The policy used in this simulation is described in Table~\ref{Table: policy_table} (CSTR, Dynamic). (\textbf{d})~The temperature of the reactant that flows into the main reactor increases throughout the trajectory. (\textbf{e})~The observations of the temperature are made more noisy. (\textbf{f})~The dynamics of the temperature in the main reactor are influenced by an external sinusoidal wave, simulating the change of temperature outside of the reactor during the day and night.}
\label{Fig: cstr_all}
\end{figure}

In the final experiment, we applied our method to the continuous stirred tank reactor (CSTR) as an industrial problem with non-linear dynamics. The temperature in the reactor has to be stabilised at a setpoint temperature by controlling the flow of coolant into the cooling jacket (Fig.~\ref{Fig: cstr_all}b). To increase the complexity of the problem, the environment setting is varied between trials and the policies receive noisy partial state observations.

The NDE did not converge after the dedicated number of generations, while being stuck at a poor fitness for the first half of the evolution (green curve). A potential reason for the limited convergence of NDEs could be that they cannot easily approximate the non-linearity required to stabilise the reactor efficiently. The validation fitness of the NDE policies is on average lower than the symbolic policies, and there are a few bad-performing policies (green dots). 

GP-S converged to a higher fitness than NDE (blue curve), and the validation fitness shows high consistency among the policies (blue dots). GP-D obtained a better fitness than GP-S during evolution (orange curve). GP-D discovered several policies that outperform GP-S and NDE in terms of validation fitness (orange dots). However, the high-performing dynamic symbolic policies were not consistently found, as some policies obtained a validation fitness significantly worse than any static policy.

Appendix Fig.~\ref{fig: variance}b presents the fitness of the best dynamic symbolic policies on the training and testing data in the continuous stirred tank reactor experiment. In this experiment, the relation between the fitness obtained during evolution and the validation fitness shows a similar negative trend for all policies. The environment has a high degree of freedom, as the initial conditions, the setpoint temperature and nine parameters are sampled randomly in every trajectory. The poor generalisation of the policies could be explained by the fact that the training data is unable to fully represent the space of possible experiment settings, and therefore the policies are tested in unfamiliar settings during validation.

The best static and dynamic policies are shown in Table~\ref{Table: policy_table} (CSTR, both Static and Dynamic). The evolved symbolic policies include non-linear terms such as logarithms and multiplications of different variables. Such complex non-linear features are more difficult to approximate for the NDE, given the number of neurons in the latent state of the NDE. Increasing the number of neurons would improve its capability to approximate any function, however the optimisation of the network would become less efficient, requiring more computation. GP evolves these non-linear functionalities easily, which highlights another advantage of our method.

\subsection{The latent variables in dynamic symbolic policies are transparent}
Across the experiments, GP was able to find successful dynamic symbolic policies. The best validation fitness obtained with each method is presented in Table~\ref{Table: fitness_table}, together with the complexity of the best policy. GP-D is generally competitive with black-box NDEs, and outperforms the NDE in 3 out of 7 experiments. The best validation fitness of GP-D is higher than GP-S for every experiment. Table~\ref{Table: fitness_table} also presents the complexity of the best policies in terms of the number of mathematical operations, variables and constants. GP-S always obtains policies with lower complexity than GP-D, which is partially accounted for by the fact that GP-D consists of three trees. The dynamic symbolic policies have an increased complexity compared to the static policies, but this complexity allows the policies to perform better in challenging experiments. Compared to the NDE, the complexity of GP-D is still significantly lower, while demonstrating competitive performance and more interpretability.

To understand the workings of a dynamic symbolic policy, we can inspect the mathematical equations and visualise the latent memory of such a policy. The best static and dynamic symbolic policies are presented in Table~\ref{Table: policy_table}. Generally, the dynamic symbolic policies consist of few variables and the equations are easily readable. The policies include sub-expressions for computing the distance to the target and state estimation from noisy observations. Nonetheless, it does not immediately become trivial to fully understand a dynamic symbolic policy purely from the equations. To this end, the latent memory can be visualised to inspect the policy's behaviour in a trajectory.

As a demonstration, the best dynamic symbolic policy discovered with GP for stabilising the CSTR is analysed for its control and transparency (Table~\ref{Table: policy_table} CSTR, Dynamic). We interpret the state equation of variable $a_1$ as a scaling factor that converges to 2.92 during the trajectory. The state equation of variable $a_2$ computes the error between the temperature and setpoint, and adds a self-connection to improve robustness to noise. The error estimate is scaled by the sum of the control, $a_2$ and a constant. The readout scales $a_2$ with the squared of $a_1$, and adds a multiplication of $a_1$ with the logarithm of the setpoint temperature. The control therefore reacts to differences between the setpoint and current temperature, scaled increasingly during the trajectory.

This policy is subjected to a set of challenging experiments on the CSTR with varying parameter settings that were not included during evolution to demonstrate its generalisability and the flexibility of the latent state. During evolution, the setpoint temperature and parameters of the environment were fixed within a trajectory, but in these example trajectories they were varied. The policy adjusted its control signal accordingly when the target changed multiple times throughout a trajectory (Fig.~\ref{Fig: cstr_all}c, green curve), when the temperature feed into the reactor changed over time (Fig.~\ref{Fig: cstr_all}d, green curve), with higher observation noise than it was evolved for (Fig.~\ref{Fig: cstr_all}e, green curve) and with an external sinusoidal influence that resembles the change of outside temperature during day and night (Fig.~\ref{Fig: cstr_all}f, green curve). In each trajectory, the policy shows a different control strategy, displaying its robustness and flexibility.

Besides robustness, these experiments also show high transparency in the control strategy to the different circumstances. The applied control signal is a mapping of the latent state that integrates the observations. As expected, the latent variable $a_1$ converges to 2.92 in the trajectories (Fig.~\ref{Fig: cstr_all}c, red curve). Latent variable $a_2$ is more expressive and shows adaptations in each experiment. In Fig.~\ref{Fig: cstr_all}c (blue curve), the latent state stabilised over time, but changed once a new setpoint temperature is required. Similarly, the latent state increased over time as the feed temperature rises (Fig.~\ref{Fig: cstr_all}d, blue curve). With higher observation noise, the latent state shows more stochasticity (Fig.~\ref{Fig: cstr_all}e, blue curve), which is also portrayed in the control signal. Finally, the latent state oscillates to compensate for the external sinusoidal influence, with a different pattern appearing after the setpoint temperature changes (Fig.~\ref{Fig: cstr_all}f, blue curve).

These experiments demonstrated that the dynamic symbolic policies can generalise to challenging environment settings that were not used during evolution. Additionally, the equations of the dynamic symbolic policy are interpretable, and the latent variables offer transparency in the use of memory. The latent variables vary in their level of expressiveness, but overall analysing the latent variables aids with understanding the functionality of a policy.

\section{Discussion}
In this paper, we presented a method to discover interpretable and high-performing white-box control systems to solve control problems. The conventional approach is to find static symbolic policies, but these are difficult to extend to real-world challenges like partial observability and volatility. To improve the robustness of control systems, the policies have to be extended with memory. Before, memory was added to symbolic policies by providing time-lagged observations~\cite{hein2018interpretable, hein2021trustworthy} or allowing the policy to explicitly store and retrieve data~\cite{koza1999genetic}. Here, we extend the policy with latent variables that are governed by differential equations. The latent variables continuously integrate observations in their dynamics, which allows for a more scalable and general approach compared to using time-lagged observations~\cite{khan2017memory}. By evolving symbolic state equations of the latent variables with genetic programming (GP), the resulting policies consist of interpretable equations and the memory is transparent.

The dynamic policies successfully solved control tasks with noisy and partial state observations, where the static policies obtained limited results. Furthermore, dynamic symbolic policies generalised better to tasks with varying environment settings than static policies. These results showed the benefits of the memory in the dynamic symbolic policies, and these policies can still be evolved efficiently with GP in terms of generations and population size. The symbolic policies were compared with a neural differential equation optimised with covariance-matrix adaptation evolution strategies as a proven black-box baseline~\cite{salimans2017evolution}. The results demonstrated that dynamic symbolic policies were evolved that generally obtain comparable performance as the black-box baseline, although competitive symbolic policies were not consistently found in every evolutionary run. 

Besides adequately solving control tasks, the dynamic symbolic policies have other advantages, such as that they consist of interpretable expressions and display transparent dynamics, as demonstrated in Fig.~\ref{Fig: cstr_all}. These features help understanding the functioning of control systems, therefore contributing to identification of undesired behaviour. Additionally, the interpretability allows us to learn new approaches to solve tasks from the policies. To further increase the interpretability in the dynamic symbolic policies, the readout could be learned as a linear numerical vector that maps the latent variables to a control signal. This way, all non-linear functionality is evolved in the latent state, and the relation between the latent variables and control signal becomes easier to understand, which is currently not always trivial (Table~\ref{Table: policy_table}). The consistency of evolution also improves, as less trees have to be optimised with GP, although another method for optimising the readout vector should be introduced.

A limitation of evolving dynamic symbolic policies with GP is that high-performing policies are not consistently discovered. Appendix Fig.~\ref{fig: variance} shows that some discovered policies displayed poor generalisation to unseen cases. In the harmonic oscillator, the limited generalisation could be partially explained by the occurrence of overfitting in some runs, related to evolving non-linear functions. In the continuous stirred tank reactor, all policies showed a decrease of performance on the validation trajectories, which is the result of the high number of free parameters in the environment. Overfitting is however a general phenomenon and not a inherent limitation of the dynamic symbolic policies or GP itself, and it shows that the introduced method is prone to overfitting when the data does not contain enough variation.

Another factor that hinders consistent evolution of high-performing policies, is the large search space that has to be traversed for the dynamic symbolic policies. The results showed that GP converges faster and more consistently when static policies were evolved under standard environment circumstances. The search space will grow even larger when the control problem contains higher-dimensional observation and control spaces. Our method demonstrates to be scalable when the environment required two control inputs (Fig.~\ref{Fig: acrobot_results}d), but this may not hold for more complex problems. Better policies can be evolved with GP when a higher population size and number of generations are selected, which also increases the computation requirements. However, previous research has focused extensively on improving the optimisation efficiency of GP, by integrating features such as optimisation of constants~\cite{topchy2001faster}, semantic reproduction~\cite{vanneschi2014survey}, population diversity control~\cite{burke2004diversity} and self-adaptation of hyperparameters~\cite{angeline1996two}.

Besides introducing the concept of symbolic policies with dynamic latent memory, this research also resulted in a general GP framework~\cite{de2025kozax}  that is not limited to optimisation of dynamic symbolic policies. First of all, other structures of symbolic policies can be configured, with the possibility to explicitly define different types of trees with unique inputs and functionality in the policy. Furthermore, the framework could also be used problems outside control, for example symbolic regression~\cite{bongard2007automated}, discovering learning rules~\cite{jordan2021evolving} or other directions within scientific discovery~\cite{wang2023scientific, vinuesa2024opportunities, rackauckas2020universal}.

We conclude that the use of genetic programming is a viable route towards learning interpretable, efficient and robust dynamic symbolic policies for solving control tasks. We hope that this adds an important tool to the practitioners' toolbox in the endeavour to create trustworthy AI systems.

\section{Acknowledgements}
This publication is part of the project ROBUST: Trustworthy AI-based Systems for Sustainable Growth with project number KICH3.L TP.20.006, which is (partly) financed by the Dutch Research Council (NWO), ASMPT, and the Dutch Ministry of Economic Affairs and Climate Policy (EZK) under the program LTP KIC 2020-2023. All content represents the opinion of the authors, which is not necessarily shared or endorsed by their respective employers and/or sponsors.

\printbibliography
\newpage

\begin{appendices}
\section{Genetic programming}
Our implementation of genetic programming follows Algorithm~\ref{alg: GP}. The hyperparameters and function sets we used in each experiment are presented in Table~\ref{Table: Hyper_params}. The choice for the population size and number of generations (see Section II-D) were chosen to balance efficiency and consistent convergence empirically. The elite percentage was set to 0.1 in every experiment. The choice of the function sets are explained in the corresponding environment sections.
\begin{figure}[h]\centering\begin{minipage}{\linewidth}\begin{algorithm}[H]
\textbf{Input} Number of generations $G$, population size $N$, elite percentage $E$, fitness function $F$
\begin{algorithmic}[1]
    \State {\em initialise} population $P$ with size $N$ (Section III-D1)
    \For{$g$ in $G$}
        \State {\em evaluate} each individual in $P$ on $F$ (Section III-D2)
         \State offspring $O$ $\longleftarrow$ \{\}
         \State append fittest $E$ of $P$ to $O$
         \While{size($O$) $<$ $N$}
         \State randomly select {\em reproduce} from \{crossover, mutation, simplification, sample\}
            \State randomly select parents $p$ from $P$ with tournament selection
            \State children $c$ = {\em reproduce}($p$) (Section III-D3)
            \State append $c$ to $O$
         \EndWhile
        \State $P$ $\longleftarrow$ $O$
    \EndFor
    \State \Return fittest individual in $P$
\end{algorithmic}\caption{Genetic programming algorithm}\label{alg: GP}
\end{algorithm}\end{minipage}\end{figure}

\begin{table}[!h]
\centering
\caption{Hyperparameters and function sets of the genetic programming algorithm used in each experiment.}\label{Table: Hyper_params}\vspace{0.2cm}
\begin{small}\centering
\begin{tabular}{lccccccc}
 \toprule
 &\multicolumn{3}{c}{\textbf{SHO}}&\multicolumn{3}{c}{\textbf{Acrobot}}&\multicolumn{1}{c}{\textbf{CSTR}}\\
    \cmidrule(lr){2-4} \cmidrule(lr){5-7} \cmidrule(lr){8-8}
 \textbf{Hyperparameter}&\textbf{Exp 1} &\textbf{Exp 2}&\textbf{Exp 3}&\textbf{Exp 1}&\textbf{Exp 2}&\textbf{Exp 3}&\textbf{Exp 1}\\
 \midrule
 Population Size      & 500 & 500 & 1000 & 500 & 500 & 500& 1000\\
 Number of Generations& 30 & 50 & 100 & 50 & 50 & 50 & 100\\
 Latent state size    & 2 & 2 & 2 & 2 & 2 & 3 &2 \\
 Function set            & \multicolumn{3}{c}{$+,-,*,/$, power}&\multicolumn{3}{c}{$+,-,*,/,$ power, sin, cos}&\multicolumn{1}{c}{$+,-,*,/,$ power, exp, log}\\
 \bottomrule
\end{tabular}
\end{small}
\end{table}

\section{Stochastic Harmonic Oscillator}
\label{app:SHO}
The dynamics of the stochastic harmonic oscillator (SHO) are given by
\begin{equation}\label{Eq:HO_dyn}
\dd\mathbf{x} = (\mathbf{Ax} + \mathbf{b}{u})\dd t + \mathbf{v}\dd{w}
\end{equation} 
with
\begin{equation}\label{Eq:HO_param}
    \mathbf{A} = \begin{bmatrix} 0 & 1\\ -\omega & -\zeta \end{bmatrix},\quad \mathbf{b} = \begin{bmatrix}  0 \\  1 \end{bmatrix},\quad
    \mathbf{v} =\begin{bmatrix} 0\\  0.05 \end{bmatrix}.
\end{equation}

Depending on the choice of $\omega$ and $\zeta$ we obtain different harmonic oscillators, reflecting different choices of the oscillator's mass, spring constant and damping.

The fitness function was chosen as the negative quadratic cost function, defined as
\begin{equation}\label{Eq:HO_fit}
    F(\mathbf{x}_{0:f},\mathbf{u}_{0:f}) 
        = - \sum_{t=0}^{f} (\mathbf{x}_t-\mathbf{x}^\ast)^\top\mathbf{Q}(\mathbf{x}_t-\mathbf{x}^\ast) - r u_t^2 
\end{equation}
with $\mathbf{Q} = \textrm{diag}(0.5, 0)$ and $r=0.5$. 

Here, $\mathbf{Q}$ punishes the policy for the distance between $\mathbf{x}$ and some target state $\mathbf{x}^\ast$ and $r$ 
penalises the control strength. The initial condition $\mathbf{x}_0$ is randomly sampled from a normal distribution with zero mean and variance of 3 and 1 for the position and velocity respectively. The target position is chosen randomly by sampling uniformly between -3 and 3 in each trial, while the target velocity is always set to zero. 

GP and the baselines are tested on the SHO in three different experiments. In the first experiment the policies interact with the standard setting of the SHO. In this experiment $\mathbf{A}$ is specified with $\omega = 1$ and $\zeta = 0$ and the state observations are noisy. In the second experiment the state of the SHO is partially observed, as the policies only observe the position while the velocity is unknown. $\mathbf{A}$ remains the same in this experiment. In the third experiment, the policies are evaluated on generalisability. In each trial $\omega$ and $\zeta$ are different, specifically $\omega \sim U(0,2)$ and $\zeta \sim U(0,1.5)$. In this experiment both the position and velocity are observed again.
In the experiments with observation noise, the parameters of the readout function are set to $\mathbf{D} = \mathbf{I}$ and $\mathbf{\Sigma} = \sigma\mathbf{I}$ with $\sigma=0.3$ (see Equation~
2). In experiments with partial state observability, $\mathbf{D}$ is adapted such that only the relevant observations are returned and the dimension of $\mathbf{\Sigma}$ is decreased to match the number of observations.

\section{Acrobot}
\label{app:acrobot}
The dynamics of the acrobot are defined as in~\cite{sutton1995generalization}.
Let $\theta_1$ and $\theta_2$ denote the angles of the first and second link, respectively. Their dynamics are given by
\begin{align}
     \ddot{\theta}_1 &= -\frac{d_2 \ddot{\theta}_2 + \phi_1}{d_1}\\
    \ddot{\theta}_2 &= \frac{u_1 + d_1^{-1} d_2 \phi_1 - m_2 l_1 lc_2 \dot{\theta}_1^2 \sin(\theta_2) - \phi_2}{m_2 l_{c2}^2 + I_2 - d_1^{-1}  d_2^2}
\end{align}
with
\begin{align}
    d_1 &= m_1 l_{c1}^2 + m_2 \left(l_1^2 + l_{c2}^2 + 2 l_1 l_{c2} \cos\left(\theta_2\right)\right) + I_1 + I_2\\
    d_2 &= m_2 \left(l_{c2}^2 + l_1 l_{c2} \cos\left(\theta_2\right)\right) + I_2\\
    \phi_1 &= -m_2 l_1 l_{c2} \dot{\theta}_2^2 \cos\left(\theta_2\right) - 2 m_2 l_1 l_{c2} \dot{\theta}_2 \dot{\theta}_1 \sin\left(\theta_2\right) \\
    &+ \left(m_1 l_{c1} + m_2 l_1\right) g \cos\left(\theta_1 - \frac{\pi}{2}\right) + \phi_2\\
    \phi_2 &= m_2 l_{c2} g \cos\left(\theta_1 + \theta_2 - \frac{\pi}{2}\right) 
\end{align}
with parameters of the system as shown in Table~\ref{table:acrobot_params}.
In the acrobot experiment that requires two control forces, the state equation of $\dot{\theta}_1$ is adapted to $$\ddot{\theta}_1 = \frac{(u_2 - d_2 \ddot{\theta}_2 - \phi_1)}{d_1}.$$
In practice, we convert this system to a stochastic first-order system
\begin{equation}
    \dd\mathbf{x}=f(\mathbf{x}, \mathbf{u})\dd t + \mathbf{V} \dd\mathbf{w}
\end{equation}

with $\mathbf{x} = (\theta_1,\theta_2,\dot{\theta}_1,\dot{\theta}_2)^\top$ where $\mathbf{V} = \textrm{diag}(0, 0, 0.05, 0.05)$
adds noise to the angle accelerations.

The fitness function is a sparse reward function with control regularisation, defined as
\begin{equation}\label{Eq:Acr_fit}
    F(\mathbf{x}_{0:f},\mathbf{u}_{0:f}) = - t_f - \sum_{t=0}^{f} \mathbf{u}_t^\top \mathbf{R}\mathbf{u}_t 
\end{equation}
where $t_f$ describes the time point at which the swing up is accomplished. The control $\mathbf{u}$ is clipped between -1 and 1, forcing policies to learn to build momentum to accomplish the swing up. The swing up is accomplished when
    $- \cos(\theta_{1}) - \cos(\theta_{1} + \theta_{2}) > x^\ast,$
where $x^\ast$ is a fixed target height in every trial. Hence, the policy must try to minimise $t_f$ and its control use. 

In every experiment, $\mathbf{R}\in \mathbb{R}^{C\times C}$ is set to $0.01 \mathbf{I}$, $x^\ast$ to 1.5 and the initial states are sampled from $U(-0.1,0.1)$. Observations of $\theta_i$ are wrapped between [-$\pi$, $\pi$] and a trial terminates when $\dot{\theta_1} > 8\pi$ or $\dot{\theta_2} > 18\pi$.
Policies are discovered to solve the acrobot in three different experiments. In the first experiment the state of the acrobot is observable with noise and control force is only applied to the second link. In the second experiment the acrobot state is made partially observable, as the policies no longer observe the angular velocities. In the third experiment, the policies can apply independent forces to both links, and the angular velocities are observed again. The observations of the acrobot include angles, therefore the sine and cosine functions are added to the function set of the GP algorithm, besides the arithmetic operators.
As for the SHO, the parameters of the readout function are set to $\mathbf{D} = \mathbf{I}$ and $\mathbf{\Sigma} = \sigma\mathbf{I}$ with $\sigma=0.3$ to add observation noise. In experiment with partial state observability, $\mathbf{D}$ is changed to exclude the unobserved variables and $\mathbf{\Sigma}$ is set to the correct dimension.

\begin{table}[t]
    \caption{Parameters of the acrobot swing up task.}\vspace{0.2cm}
    \label{table:acrobot_params}
    \centering
    \begin{tabular}{llll}
 \toprule
 \textbf{Parameter}& \textbf{Unit}&\textbf{Description} &\textbf{Value or Range}\\
    \midrule
    $\theta_i$ & \unit{\radian} & Angle of a link & $\theta_i(0) \sim U(-0.1,0.1)$\\
    $\dot{\theta}_i$ & \unit{\radian\per\second} & Angular velocity of a link & $\dot{\theta}_i(0) \sim U(-0.1,0.1)$\\
    $l_i$ & \unit{\meter} & Length of a link & 1.0\\
    $m_i$ & \unit{\kilogram} & Mass of a link & 1.0\\
    $l_{ci}$ & \unit{\meter} & Position of the center of mass of a link & 0.5\\
    $I_i$ & \unit{\kilogram \squared \meter} & Moment of inertia of a link & 1.0\\
    $g$ & \unit{\meter\per\square\second} & Gravity & \num{9.81}\\
    \hline
    \end{tabular}
\end{table}

\begin{table}[!h]
\centering
\caption{Parameters of the continuous stirred tank reactor. Each parameter is either set to a value, a function or sampled uniformly given a range.}\label{Table: CSTR_table}\vspace{0.2cm}
\begin{tabular}{llll }
 \toprule
 \textbf{Parameter}& \textbf{Unit}&\textbf{Description} &\textbf{Value or Range}\\
 \midrule
 $c$ & \unit{\mole\per\liter} & Reactant concentration    &$c(0) \sim U(0.5,1.0)$\\
 $T_r$& \unit{\K}&   Reactor temperature  & $T(0) \sim U(350,375)$   \\
 $T_c$& \unit{\K}&Cooling jacket temperature & $T_c(0) \sim U(275,300)$\\
 $T_r^\ast$&\unit{\K} & Setpoint temperature & \{\num{400}, \num{500}\}\\
 $q_r$  &  \unit{\liter\per\minute}&Feed flowrate of reactant into reactor & \{\num{75}, \num{125}\}\\
 $q_c$& \unit{\liter\per\minute}&   Feed flowrate of coolant into cooling jacket  & Controlled parameter\\
 $c_f$&  \unit{\mole\per\liter}& Feed concentration of reactant into reactor  & \num{1}   \\
 $T_f$&  \unit{\K}& Feed temperature of reactant into reactor  & \{\num{300}, \num{350}\}\\
 $T_{cf}$& \unit{\K}& Feed temperature of coolant into cooling jacket  & \{\num{250}, \num{300}\}\\
 $V_r$& \unit{\liter}& Volume of reactor  & \{\num{75}, \num{150}\}\\
 $V_c$& \unit{\liter}& Volume of cooling jacket  & \{\num{10}, \num{30}\}\\
 $\Delta H$& \unit{\joule\per \mole}& Enthalpy of reaction  & \{\num{-55000}, \num{-45000}\}\\
 $\rho$& \unit{\gram\per\liter}&Density  & \num{1000}\\
 $C_p$& \unit{\joule\per\gram\per\K}& Heat capacity & \{\num{0.2}, \num{0.35}\}\\
 $UA$& \unit{\joule\per\minute\per\K}& Heat transfer coefficient  & \{\num{25000}, \num{75000}\}\\
 $k_0$& \unit{1\per\minute}& Arrhenius pre-exponential  & $\num{7.2e10}$\\
 $E$& \unit{\joule\per\mole}& Activation energy  & \num{72750}\\
 $R$& \unit{\joule\per\mole\per\K}& Gas constant  & \num{8.314}\\
 $k(T_r)$& \unit{1\per\minute}& Reaction rate  & Function of temperature\\
 \bottomrule
\end{tabular}
\end{table}

\section{Continuous stirred tank reactor} \label{app_b}
The dynamics of the CSTR are defined as in~\cite{antonelli2003continuous}, extended with a Wiener process to incorporate stochasticity. Let $\mathbf{x} = (c, T_r, T_c)^\top$. 
The CSTR is defined as a stochastic first-order system 
\begin{equation}
    \dd \mathbf{x}=f(\mathbf{x}, \mathbf{u})\dd t + \mathbf{V} \dd\mathbf{w}
\end{equation}
with $\mathbf{V} = \textrm{diag}(0.025, 3, 3)$.
The state equation is defined as \begin{equation}
f(\mathbf{x}, \mathbf{u}) = 
        [f_1(\mathbf{x}, \mathbf{u}), 
        f_2(\mathbf{x}, \mathbf{u}), 
        f_3(\mathbf{x}, \mathbf{u})]^\top\end{equation}
with
\begin{align}
    f_1(\mathbf{x}, \mathbf{u})&=\frac{q_r}{V_r} \left(c_f-c\right)+k(T_r)c\\
    f_2(\mathbf{x}, \mathbf{u})&=\frac{q_r}{V_r}\left(T_f-T_r\right)+\frac{-\Delta H}{\rho C_p} k(T_r) c + \frac{UA}{\rho C_p V_r} \left(T_c-T_r\right)\\
    f_3(\mathbf{x}, \mathbf{u})&=\frac{q_c}{V_c}\left(T_{cf}-T_c\right) + \frac{UA}{\rho C_p V_r} \left(T_r-T_c\right).
\end{align} 
The descriptions of each parameter in the CSTR dynamics are shown in Table \ref{Table: CSTR_table}. Table \ref{Table: CSTR_table} also contains the value, initial condition or sample range for each parameter. The variables that are indicated with a range are uniformly sampled within this range in every trajectory. The initial conditions of $c$, $T_r$ and $T_c$ are uniformly sampled from the given distribution. The control signal is clipped between 0 and 300. The reaction rate is defined as $k(T_r)=k_0 \exp\left(-{E}/{RT_r}\right)$. 

The fitness function is chosen as the negative quadratic cost function, defined as
\begin{equation}\label{Eq:CSTR_fit}
    F(\mathbf{x}_{0:f},\mathbf{u}_{0:f}) = - \sum_{t=0}^{f} (\mathbf{x}_t-\mathbf{x}^\ast)^\top\mathbf{Q}(\mathbf{x}_t-\mathbf{x}^\ast) - r u_t^2  
\end{equation}
where
\begin{equation}\label{Eq:CSTR_param}
    \mathbf{Q} = \begin{pmatrix} 0 & 0 & 0\\ 0 & 0.01 & 0 \\ 0 & 0 & 0\end{pmatrix} ,\quad r = 0.0001
     ,\quad \mathbf{x}^\ast = \begin{pmatrix} 0 \\ T_r^\ast \\ 0\end{pmatrix}.
\end{equation}
The policy has to minimise the distance between the temperature in the reactor and the setpoint temperature $T_r^\ast$, while also minimising the control, i.e. the coolant flowrate.

Only a single experiment is performed regarding the CSTR. This experiment tests policies for their generalisability to an industrial environment. The parameters of the CSTR are different in each trial, to simulate as if the policy is operating in multiple scenarios. In total, eight parameters vary between trials (Table~III). 
Additionally, the concentration in the reactor is not observed, and the observations of the temperature in the reactor and cooling jacket are noisy. The function set of the GP algorithm is extended to include the exponential and logarithmic functions to aid the policies in dealing with the values of the temperatures, which typically range between 200 and 600. To generate the observations, $\mathbf{D}$ is defined such that the concentration is excluded from the observations and $\mathbf{\Sigma}=\sigma\mathbf{I}$ with $\sigma=7.5$.

\newpage
\section{Comparison with time-lagged policies}
\begin{table}[h]
\centering
\caption{The time-lagged static and dynamic symbolic policies discovered with genetic programming for controlling the partially observable harmonic oscillator with various levels of noise ($\sigma$). $u_j$ represents the function that outputs the control signal. In the time-lagged policies, the second index in the observations $y_i,t$ indicates the delay. In the dynamic policies, the functions of $a_i$ represent the state equations of the latent variables. When a latent variable is not referred to in any of the other expressions, it is labelled as inactive.}
\vspace{0.2cm}
 \begin{small}
 \centering
\begin{tabular}{llll}
\toprule
 \textbf{$\sigma$}& \textbf{Static policy}& \textbf{Time-lagged policy} &\textbf{Dynamic policy}\\
 \toprule
 0.0 & $\begin{array}{lll}u_1 &=& 0.25y_1 + 0.48 x^* \\&&- 0.10\end{array}$ & $\begin{array}{lll}u_1 &=& 35.44 (y_{1,-1} - y_{1,-2}) \\&&+ x^*\end{array}$ & $\begin{array}{lll}u_1 &=& a_1 + x^* \\ \dot{a}_1 &=& -2a_1 + a2 - x^* - 0.18\\ \dot{a}_2 &=& y_1 - 2u_1 + x^*\end{array}$\\ \midrule
 0.25 & $\begin{array}{lll}u_1 &=& 0.25y_1 + 0.49 x^*\end{array}$ & $\begin{array}{lll}u_1 &=& 3y_{1,-1} - y_{1,-2} - 2y_{1,-3} \\&&+ 0.97x^*\end{array}$ & $\begin{array}{lll}u_1 &=& 1.40a_1 + 0.90a_2 + x^* \\ \dot{a}_1 &=& a_2 + y_1 + x^* \\ \dot{a}_2 &=& - 1.97u_1 + y_1 + x^* \end{array}$\\ \midrule
 0.5 & $\begin{array}{lll}u_1 &=& 0.26y_1 + 0.47 x^*\\&& - 0.05\end{array}$ & $\begin{array}{lll}u_1 &=& 2y_{1,-1} - 2y_{1,-3} + x^*\end{array}$ & $\begin{array}{lll}u_1 &=& a_1 + x^* \\ \dot{a}_1 &=& - 0.72a_1 + 0.57y_1 - 0.57x^*\\ \dot{a}_2 &=& 0.14\, \text{(Inactive)} \end{array}$\\ \midrule
 0.75 & $\begin{array}{lll}u_1 &=& 0.23y_1 + 0.52 x^*\end{array}$ & $\begin{array}{lll}u_1 &=& 1.18y_{1,-1} - 0.91y_{1,-3} \\&&+ 0.62x^*\end{array}$ & $\begin{array}{lll}u_1 &=& a_2 + x^* \\ \dot{a}_1 &=& -1.53(a_1 + y_1) \\ \dot{a}_2 &=& -a_1 - 0.43a_2 - u_1 - 0.01 \end{array}$\\ \midrule
 1.0 & $\begin{array}{lll}u_1 &=& 0.25y_1 + 0.47 x^* \\&&- 0.05\end{array}$ & $\begin{array}{lll}u_1 &=& y_{1,-1} - 0.77y_{1,-3} + 0.66x^*\\&& - 0.06 \end{array}$ & $\begin{array}{lll}u_1 &=& 2.53a_1 + 3.53a_2 \\ \dot{a}_1 &=& - 2a_1 - y_1 + x^*\\ \dot{a}_2 &=& -a_1 -u_1 + x^* \end{array}$\\
 \bottomrule
\end{tabular}
\end{small}
\label{Table: lag_policy_table}
\end{table}

\newpage
\section{Variance Analysis}
\begin{figure}[h]
    \centering
    \includegraphics[width=0.75\linewidth]{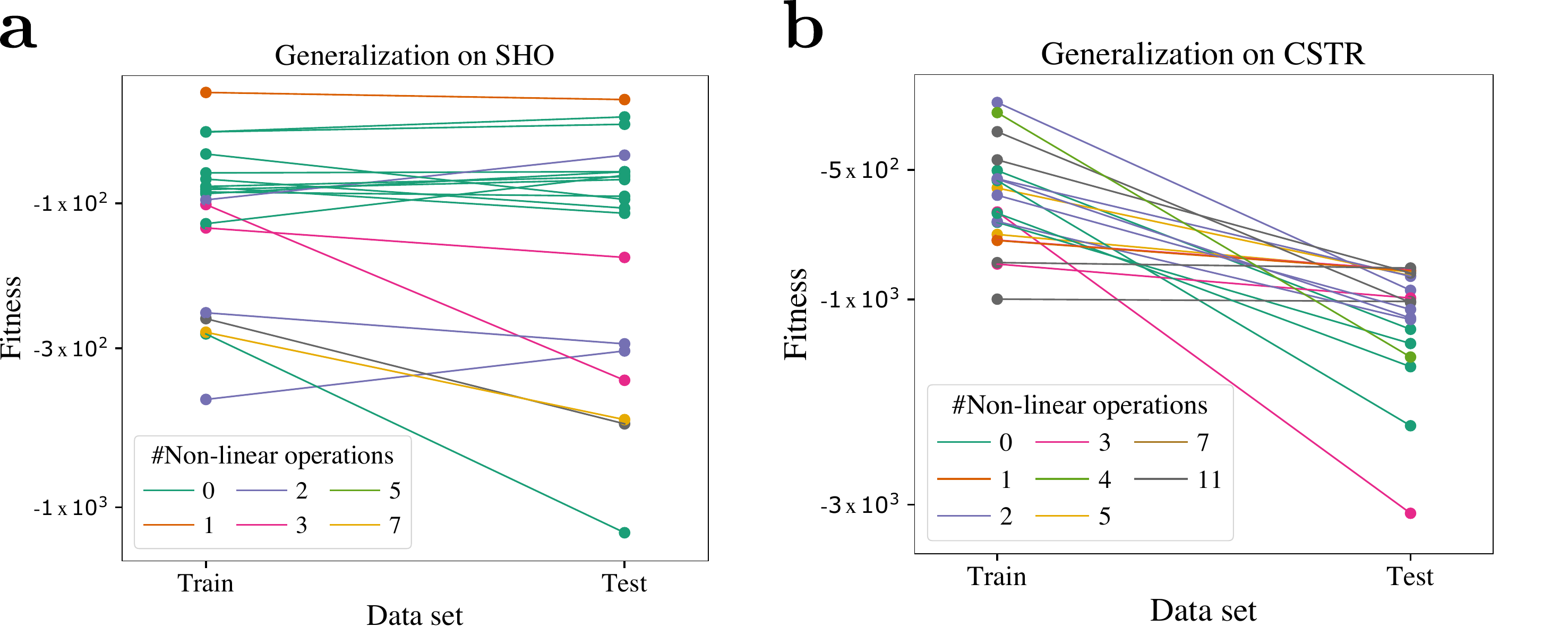}
    \caption{\textbf{Comparison of fitness on training and testing data.} (\textbf{a})~The fitness obtained by the best dynamic symbolic policy found during evolution on the stochastic harmonic oscillator (SHO) task with varying environment parameters is plotted. Additionally, the fitness of these policies obtained on unseen test data. The colour indicates the number of non-linear operations, counting multiplications between pairs of variables and the occurrences of divisions and powers. (\textbf{b})~The train and test fitness of the best dynamic symbolic policies from the continuous stirred tank reactor (CSTR) experiment. The non-linear operations can also include exponentials and logarithms here.}
    \label{fig: variance}
\end{figure}
\end{appendices}
\end{document}